\documentclass{IOS-Book-Article}

\usepackage{mathptmx}
\usepackage{soul}\setuldepth{article}
\usepackage{booktabs}
\usepackage{microtype}
\usepackage{tikz}
\usepackage{fancyvrb}
\usepackage{graphicx}
\usepackage{subfigure}
\usepackage{multirow}
\def\hb{\hbox to 11.5 cm{}}

\begin{document}

\pagestyle{headings}
\def\thepage{}
\begin{frontmatter}              

\title{A Question-Answering Approach to\\Evaluating Legal Summaries}

\markboth{}{December 2023\hb}

\author[A, B]{Huihui Xu
\thanks{Huihui Xu is the corresponding author and can be contacted via email: huihui.xu@pitt.edu.}},
\author[A, B, C]{Kevin Ashley}

\address[A]{Intelligent Systems Program, University of Pittsburgh}
\address[B]{ Learning Research and Development Center, University of Pittsburgh}
\address[C]{School of Law, University of Pittsburgh}

\begin{abstract}
Traditional evaluation metrics like \verb|ROUGE| compare lexical overlap between the reference and generated summaries without taking argumentative structure into account, which is important for legal summaries. In this paper, we propose a novel legal summarization evaluation framework that utilizes \textbf{GPT-4} to generate a set of question-answer pairs that cover main points and information in the reference summary. GPT-4 is then used to produce answers based on the generated summary for the questions from the reference summary. Finally, GPT-4 grades the answers from the reference summary and the generated summary. We examined the correlation between GPT-4 grading and human grading. The results suggest that this question-answering approach with GPT-4 can be a useful tool for gauging the quality of the summary.
\end{abstract}

\end{frontmatter}
\markboth{December 2023\hb}{December 2023\hb}

\section{Introduction}
Readers need summaries to convey a rough idea of what a case is about and why it is important. This enables users to connect a case to their personal needs and to decide whether to read the case decision. As a result, the quality of legal summaries is important. Commonly used summary evaluation metrics such as \verb|ROUGE| scores \cite{lin2004rouge}  focus primarily on surface-level aspects like word overlap and grammatical correctness. These metrics do not consider factors such as contextual understanding or the alignment of the summary with the reader's specific goals or preferences. 

In this work, we propose a novel method to evaluate the quality of a legal summary by leveraging automated question-answering while incorporating legal argumentative structure. The argument structure comprises  three elements: \textbf{Issue} -- legal question that a court addressed in the case; \textbf{Reason} -- elaboration of why the court reached the conclusion; and 
\textbf{Conclusion} -- the court's final decision regarding the issue. Our method consists of three steps: (1) Given a reference legal summary, a question-answer generation model (GPT-4) produces a set of question-answer pairs based on the legal argumentative structure of the reference summary. (2) Then we use a question-answering model (GPT-4) to answer the questions
from the reference summary based on the text of the generated summary. (3)  Finally, GPT-4 compares the answers in step (1) from the reference summary with the answers in step (2) from the generated summary and assigns grades based on the similarity. Code is available at \url{https://github.com/JoyceXu02/QA_evaluation}.

\section{Related Work}
 The Stanford Question Answering Dataset (SQuAD) introduced in \cite{rajpurkar2016squad} is  useful for training and evaluating question-answering (QA) systems. It contains questions posed on a set of Wikipedia articles, where the answer to each question is a segment of text from the corresponding passage. In this work, we are inspired by the idea of assessing the quality of summaries by asking questions about them: if a good summary retains all crucial information, then it should be able to answer questions about the original content accurately. Researchers in \cite{eyal2019question, scialom2019answers} proposed and concluded that human evaluators prefer QA-based metrics for evaluating abstractive summaries. Taking the QA-based evaluation research methods even further, \cite{durmus2020feqa} tackled the unfaithfulness of neural abstractive summarization by proposing QA-based automatic metrics, FEQA. Our work inherits the idea of using a QA-based method to evaluate the summary quality while taking legal argument into account.

Basing an evaluator  on a large language model (LLM) has become increasingly popular lately. The authors of \cite{fu2023gptscore} proposed an evaluation framework, GPTScore, with generative pre-training models like GPT-3. They  suggest that higher-quality text is more likely to be generated by following a given instruction in a given context. Researchers in \cite{wang2023chatgpt, liu2303g} present a preliminary study of using LLMs as a Natural Language Generation (NLG) evaluator. Their LLM evaluation achieved a new state-of-the-art correlation in the summary evaluation task.

Few prior projects apply a question-answering   approach to evaluation in a legal context. Our approach leverages recent progress in large language models while taking legal argumentative structure into account. Before the bloom of LLMs, an QA approach relied on corresponding curated datasets \cite{anantha2020open, adlakha2022topiocqa}. Our approach does not require a specific question-answering dataset and generates questions and answers automatically.

\section{Methodology}
\subsection{Experimental Design}
We utilize GPT-4 to generate question-answer pairs, incorporating the example prompt illustrated in Figure \ref{fig:qa_prompt}. This enhanced prompt enables us to generate not only question-answer pairs but also the corresponding question types. Subsequently, we utilized these questions as prompts to the model for predicting responses based on model-generated summaries. The prompt used for the prediction is shown in Figure \ref{fig:pred_prompt}. The prompt for GPT-4 to evaluate the answers is shown in Figure \ref{fig:eval_prompt}. We set the temperature to 0 for the GPT-4 generation part to get the most deterministic results. When human evaluators assess the quality of automatically generated summaries, they use previously generated question-answer pairs as a guide or reference. These pairs help the evaluators know what to look for and how to judge the summary's quality.

Throughout our research, we experimented with three models for generating summaries: Longformer Encoder-Decoder (LED) \cite{beltagy2020longformer}, BART \cite{lewis2020bart}, and GPT-4. LED and BART require fine-tuning in order to generate reasonable summaries while GPT-4 can generate summaries in a zero-shot setting. 

\subsection{Data}
We  developed a type system to annotate Canadian legal case summaries\cite{Xu2022MultigranularityAM, xu2021accounting}. This type system includes three key components: Issue, Reason, and Conclusion. The dataset initially consisted of 1,049 annotated summaries along with their corresponding full-case decisions. We used the same dataset to support this work.

We used 90\% of the data for fine-tuning LED and BART models. The remaining 10\% of the data was for testing purposes. GPT-4 was used to generate summaries for this 10\% subset of data without fine-tuning.  Considering the cost associated with GPT-4 and human evaluation, however, we opted to leverage our QA approach to evaluate 10 summaries generated by each model.

\begin{figure}[h]
\footnotesize
\begin{Verbatim}[frame=single,commandchars=\\\{\}]
Act like a legal professional and read the following legal text. 
Use \textcolor{red}{Issue}, \textcolor{blue}{Reason}, \textcolor{teal}{Conclusion} sentences to generate question-answer pairs [...] 

List those generated questions and answers in the following format: 

Question: {\string{\string{generated_qustion\string}\string}}
Type:{\string{\string{question_type\string}\string}}
Answer:{\string{\string{generated_answer\string}\string}}
\end{Verbatim}
\caption{Prompt template for generating question-answer pairs based on annotated sentences.}
\label{fig:qa_prompt}
\end{figure}

\begin{figure}[t]
\footnotesize
\begin{Verbatim}[frame=single,commandchars=\\\{\}]
Answer the {\string{\string{type\string}\string}} question based on the context
Context:{\string{\string{context\string}\string}}
Question:{\string{\string{question\string}\string}}
Answer: 
\end{Verbatim}
\caption{Prompt template for predicting answers based on the model-generated summary. }
\label{fig:pred_prompt}
\end{figure}

\begin{figure}[t]
\footnotesize
\begin{Verbatim}[frame=single,commandchars=\\\{\}]
You are a legal expert to judge the answers to questions. 

You are judging the following question:
{\string{\string{qustion\string}\string}}
The real answer is: 
{\string{\string{real_answer\string}\string}}
You are grading the following predicted answer: 
{\string{\string{predicted_answer\string}\string}}

What grade do you give on a scale from 0 to 10, where 0 is the lowest (can't find 
the answer) and 10 is the highest (very close to the real answer). 

Finally, give the explanation of the grade
Explanation: {\string{\string{explanation\string}\string}}
\end{Verbatim}

\caption{Prompt template to evaluate the predicted answer with the real answer. }
\label{fig:eval_prompt}
\end{figure}

\section{Results and Discussion}

There are 48 question-answer pairs for 10 cases. A human evaluator assessed whether the generated question-answer pairs adequately captured the necessary information and were  addressed correctly. The evaluation options were limited to ``YES" and ``NO". Based on the results, 42 out of the 48 questions accurately captured the required information, while all 48 answers were correct and appropriately addressed the questions. Table ~\ref{tab:qa_example1} shows an example of GPT-4-generated QAs. This example shows that GPT-4 can generate coherent and contextually relevant answers to specific types of questions. Those QAs  served as ground truth when comparing to the predicted grading. 

\begin{table}[th!]
  \centering
  \caption{GPT-4 generated QA examples. The questions in red pertain to an issue; the one in blue is focused on the reason, and the question in teal color aims at drawing the conclusion.}
  \label{tab:qa_example1}
  \begin{tabular}{|p{0.3\linewidth}|p{0.2\linewidth}|p{0.3\linewidth}|}
    \hline
   \textbf{Summary} & \textbf{Question} & \textbf{Answer} \\
    \hline
    \multirow{3}{3.5cm}{Warrant issued to search a dwelling house for weapons allegedly used in an attempted armed robbery. The affidavit in support referred to an unknown informant. Judge applied the test that the justice of the peace `must be satisfied on reasonable grounds.' Substantial compliance found and warrant upheld.} & \textcolor{red}{What was the warrant issued for?} & 	The warrant was issued to search a dwelling house for weapons allegedly used in an attempted armed robbery. \\
   & \textcolor{blue}{What test did the judge apply to determine the validity of the warrant?} & The judge applied the test that the justice of the peace'must be satisfied on reasonable grounds.' \\
    & \textcolor{teal}{What was the conclusion of the case?} & Substantial compliance was found and the warrant was upheld. \\
    \hline
  \end{tabular}
\end{table}

The prompt in figure \ref{fig:pred_prompt} is for predicting answers based on the previous generated questions and generated summaries. Figure \ref{fig:eval_prompt} shows the prompt we used for grading the predicted answer against the ground truth. We converted GPT-4 grades (0-10 scale) into binary by setting a threshold. We recognize that the numerical meaning of a generated score might be different from human perception. To avoid misinterpretation, we show results at thresholds 5 and 6 for sensitivity analysis. The grades equal and above the threshold map to ``YES" and grades below the threshold map to ``NO". Furthermore, the human evaluator assessed whether the generated answer correctly addresses the given question in relation to the model-generated summary. The evaluation options were also limited to ``YES" and ``NO". During the assessment, the human evaluator found some of the answers were legally correct, but went well beyond the information provided in the generated summary. The evaluator also found that some answers contained hallucinations. 

Table \ref{human_eval_table} shows the two types of correlation between GPT-4 evaluation grade and human evaluation at two set of thresholds (5 and 6). Pearson correlation measures the linear relationship between GPT-4 and human evaluations, while Spearman correlation measures the monotonic relationship between the two types of evaluations. Those metrics measure two aspects of the relationship between two variables. Here, we focus solely on the results at the threshold of 5 which closely align with those at threshold 6. Automatic evaluation of BART-generated summaries has highest correlation with human evaluation on Issue types of answers with Spearman (0.72) correlations. LED-generated summaries have the highest correlation on Reasons (0.43 in Pearson and 0.31 in Spearman) and highest Pearson correlation on Issue (0.69).  GPT-4 generated summaries have the highest correlation on Conclusions (0.57 in both Pearson and Spearman). In terms of Reason types of answers, we notice that the automatic evaluation has a negative correlation with the human evaluation on BART generated and GPT-4 generated summaries in both Pearson and Spearman correlation. 

Overall, LED exhibits robust linear (0.87) and monotonic (0.84) relationships. The correlation results of BART suggest that the Spearman relationship is stronger (0.54). GPT-4 has a stronger Pearson correlation (0.50). 


\section{Conclusion and Future Work}
In conclusion, our QA approach for evaluation tends to align with human judgements on  Issue and Reason type answers. It could be useful assessing the quality of summarization systems.  The strength of this correlation, however, can vary depending on the model and specific aspects of the evaluated summary. As a result, while the method offers valuable insights, it is best used along with other metrics for a comprehensive assessment.

\begin{table}[t!]
\caption{The correlation between the GPT-4 evaluation grade and the human evaluation.  IRC is short for Issue, Reason and Conclusion. The IRC correlation is computed across all sentences related to issues, reasons, and conclusions without categorizing or grouping them by these types. Number in parentheses is the chosen threshold.}
\label{human_eval_table}
\vskip 0.15in
\begin{center}
\begin{small}
\begin{sc}
\begin{tabular}{p{1cm}p{1.4cm}cccc}
\toprule

Model &Type &Pearson(5) &Spearman(5)&Pearson(6) &Spearman(6)  \\
\midrule
\multirow{3}{*}{BART} & Issue &0.67 &0.72&0.67 &0.72\\
                      & Reason &-0.07 &-0.17&-0.07 &-0.17\\
                      & Conclusion &0.29 &0.29&0.29 &0.29\\
\multirow{3}{*}{LED} & Issue & 0.69 &0.52& 0.69 &0.52\\
                      & Reason &0.43 &0.31&0.45 &0.36\\
                      & Conclusion &0.12 &0.12&0.12 &0.12\\
\multirow{2}{*}{GPT-4} & Issue &0.56 &0.56&0.56 &0.56\\
                      & Reason &-0.09& -0.20&0&-0.11\\
                      & Conclusion &0.57&0.57 &0.57&0.57 \\
BART &IRC &0.51 &0.54 &0.51 &0.54 \\ 
LED  &IRC &0.87 &0.84 &0.88 &0.85\\
GPT-4 &IRC &0.50 &0.48 &0.52 &0.48 \\
\bottomrule
\end{tabular}
\end{sc}
\end{small}
\end{center}
\end{table}

While we show that GPT-4 achieves reasonable correlation with human evaluation of summaries, there are limitations that provide directions for future work: (1) Since GPT-4’s performance as an evaluation metric is sensitive
to the construction of prompts,  we plan to explore various prompts to achieve better performance.
(2) We need to scale up the experiment to show more robust comparison results. (3) Quality control of model generation is necessary, especially when the input context increases in length and structural complexity. We will further explore open-source models to calibrate the output and ensure consistency.

\section*{Acknowledgements}

This work has been supported by grants from the Autonomy through Cyberjustice Technologies Research Partnership at the University of Montreal Cyberjustice Laboratory and
the National Science Foundation, grant no. 2040490, FAI: Using AI to Increase Fairness by Improving Access to Justice. The Canadian Legal Information Institute provided
the corpus of paired legal cases and summaries. This work was supported in part by the University of Pittsburgh Center for Research Computing through the resources provided. Specifically, this work used the H2P cluster, which is supported by NSF award number OAC-21176.

\bibliographystyle{vancouver}
\bibliography{ios-bibliography.bib}
\end{document}